\documentclass[conference]{IEEEtran}   	% use "amsart" instead of "article" for AMSLaTeX format
\usepackage{geometry}                		% See geometry.pdf to learn the layout options. There are lots.
\usepackage[parfill]{parskip}    		        % Activate to begin paragraphs with an empty line rather than an indent
\usepackage{graphicx}				% Use pdf, png, jpg, or eps§ with pdflatex; use eps in DVI mode
\usepackage{amssymb}				% !TEX TS-program = latex   pdflatexmk
\usepackage{amsmath}
\usepackage{textcomp}
\usepackage{breqn}
\usepackage{bbm}
\usepackage{tabularx}
\usepackage{latexsym}
\usepackage{subcaption}
\usepackage{gensymb}
\usepackage{amsmath}
\usepackage{fancyhdr}
\usepackage{eucal}
\usepackage{float}
\usepackage{parskip}
\usepackage{textcomp}

\title{Chapter 3}  
\usepackage{geometry}                		% See geometry.pdf to learn the layout options. There are lots.
\usepackage{tabularx}
\usepackage{stmaryrd}
\usepackage{wasysym} 
\usepackage{xurl}
\usepackage{multirow}
\usepackage[table]{xcolor} 
\newcommand{\code}[1]{\texttt{#1}}

\usepackage{microtype}
\usepackage{yfonts}
\usepackage{listings} 
\usepackage{multirow}
\usepackage{xcolor}
\usepackage{pdfpages}
\usepackage{graphicx} 
\usepackage[utf8]{inputenc}
\usepackage [english]{babel}
\usepackage [autostyle, english = american]{csquotes}
\usepackage[backend=biber]{biblatex}
\graphicspath{{./Images/}} 
\usepackage{mathtools}
\usepackage{amsmath}
\usepackage{fancyhdr} 
\usepackage{titlesec}
\usepackage{upgreek} 
\usepackage{amsmath} 
\usepackage{amssymb}

\PassOptionsToPackage{hyphens, spaces, obeyspaces}{url} 
\usepackage[final=true,backref=true,colorlinks=true,linkcolor=blue, urlcolor=blue, citecolor=blue, breaklinks=true]{hyperref}
\hypersetup{
    pagebackref=true,
    colorlinks=true,
    citecolor=blue,
    linkcolor=blue,
    filecolor=magenta,      
    urlcolor=blue,
}
\hypersetup{breaklinks=true}
\usepackage{breakurl}
\def\xHyphenate#1#2\wholeString {\if#1$%
    \else\transform{#1}%
    \takeTheRest#2\ofTheString\fi}

\def\takeTheRest#1\ofTheString\fi
{\fi \xHyphenate#1\wholeString}

\def\transform#1{\url{#1}\hskip 0pt plus 1pt}

\definecolor{dkgreen}{rgb}{0,0.6,0}
\definecolor{gray}{rgb}{0.5,0.5,0.5}
\definecolor{mauve}{rgb}{0.58,0,0.82}
\definecolor{lightsilver}{rgb}{0.96,0.96,0.98} 
\definecolor{blond}{rgb}{1, 0.98, 0.8}
\lstdefinestyle{R} {
backgroundcolor=\color{blond},
language=Python,
basicstyle=\tiny,
} 
\titleformat{\chapter}[display]
{\normalfont\Huge\bfseries\raggedright}{\chaptertitlename\ 3}
{15pt}{\Huge}
 
\author{Justin London}
\date{December 20, 2024}

\begin{document}

%\maketitle{Design of an Efficient Three-Level Buck-Boost Converter in PSIM} \\
\title{Fourier Optics and Deep Learning Methods for Fast 3D Reconstruction in Digital Holography}
%{\footnotesize \textsuperscript{*}Note: Sub-titles are not captured for https://ieeexplore.ieee.org  and
\author{\IEEEauthorblockN{Justin London}
\IEEEauthorblockA{\textit{Department of Electrical Engineering and Computer Science} \\
\textit{University of North Dakota}\\
Grand Forks, North Dakota USA \\
justin.london@und.edu}
}
\maketitle

\begin{abstract}
    Computer-generated holography (CGH) is a promising method that modulates user-defined waveforms with digital holograms.  An efficient and fast pipeline framework is proposed to synthesize CGH using initial point cloud and MRI data. This input data is reconstructed into volumetric objects that are then input into non-convex Fourier optics optimization algorithms for phase-only hologram (POH) and complex-hologram (CH) generation using alternating projection, SGD, and quasi-Netwton methods.  Comparison of reconstruction performance of these algorithms as measured by MSE, RMSE, and PSNR is analyzed as well as to HoloNet deep learning CGH.  Performance metrics are shown to be improved by using 2D median filtering to remove artifacts and speckled noise during optimization.
\end{abstract}


\maketitle

\section{Introduction}
 Holography is an interferometric imaging method to generate a hologram, a recording of an interference pattern that can reproduce a 3D light field using diffraction. An object distribution is not imaged directly but reconstructed from an inference pattern formed by the superposition of waves scattered by the object and a reference wave \cite{Mustafi:2023}.  The object wave is diffracted light and the reference wave is illuminated light used to create an interference pattern with the diffracted light reflected from the object being recorded.   
 
 Diffracted waves of the object are superimposed in a manner to cause interference to make measurements or generate images, but that does not require an imaging lens and can be used to recover a wavefront diffracted from an object.  The hologram can be displayed on a spatial light modulator (SLM) allowing the 3D image to be reproduced in real space.  Thus since hologram displays can successfully reproduce the wavefronts of 3D objects they are suitable as 3D displays \cite{Shimobaba:2022}.   
%Common 3D imaging technologies include light field imaging [7,8], time-of-flight imaging [9], structured light imaging [10,11], etc. 3D display can be realized by 3D holographic display [12–16], binocular vision display [17], volumetric 3D display [18], light field display [19,20], etc. 3D holographic display technology realizes the reconstruction of 3D scene by using recorded wavefront information. 
The wave interference of light enables both amplitude and phase information of an object to be stored and reconstructed \cite{Gopakumar:2024}.   A three-dimensional (3D) image can be reconstructed by utilization of the theory of the diffraction of light.   In digital holography (DH), the reconstruction process is achieved via computer in contrast to the conventional use of photochemical processes \cite{Palacios:2007}.  DH is a modern optic application based on the theory of diffraction and image formation that use an image sensor to capture a hologram of real macroscale objects or microscopic cells \cite{Shimobaba:2022}. 

Holography has a wide range of applications in medicine, microscopy, the military, weather forecasting, AR/VR, digital art, and security.  Holographic technology has commercial uses.  For instance, it can be used for quality control and fracture testing. AR holograms can also be used to display interactive directions during equipment assembly.   There have been various advances in holography as technology has shifted to digital computer-generated holography (CGH) even as an investigative tool for photochemical processes. 

While various reconstruction methods to generate 3D holograms from volumetric objects and vice versa have been proposed \cite{Birdi:2020} for CGH, little research progress has been toward development of fast real-time 3D hologram reconstruction processing using point clouds and improvement in reconstruction accuracy.  
%to volumetric objects and from volumetric objects to 3D holograms. measuring the precision of the reconstruction. 
We propose a methodology for fast reconstruction of 3D holograms point clouds and 2D images and demonstrate that we can greatly improve 3D reconstruction accuracy using image filtering/sharpening during the reconstruction process.  This can advance use of CGH in applications like microscopy where reconstruction accuracy of cellular and tissue details is essential.

\section{Motivation}

Reproduced images from a hologram can be obtained numerically from diffraction computations to obtain the complex amplitude of object light, aberration corrections in the optical system, and phase unwrapping from encoding.  An SLM can only modulate amplitude or phase.  Therefore, it is necessary to encode the complex-valued hologram (CH) into amplitude or phase-only holograms (POH). The encoded hologram can be displayed on the SLM and the 3D image can be observed through the optical system \cite{Shimobaba:2022}.  DH has several advantages over conventional microscopy holography including improved focal depth, ability to generate 3D images, and phase contrast images \cite{Palacios:2007}.  
%Holography has important applications in microscopy, astronomy, entertainment and medicine.

In general, 3D CGH is implemented by encoding the phase of a coherent wavefront in the Fourier space (pupil plane) of an imaging system with a spatial light modulator (SLM). \cite{Zhang:2017} However, computing the 2D Fourier phase mask that generates a specific 3D intensity is usually an ill-posed or non-convex optimization problem with non-unique solutions in a high-dimensional space (millions of pixels) that cannot be fully explored. Most 3D intensity patterns are limited to the coherent diffraction limit set by the numerical aperture (NA), the limited degrees of freedom (DOF) for patterning, and the fact that intensity must be conserved at each depth plane \cite{Zhang:2017}.  Furthermore, there are hardware limitations such as the SLM dynamic range, pixel pitch, and pixel count.  Thus, 3D CGH solutions are approximate where there is no exact solution. However, the proposed pipeline for CGH simulates an SLM overcoming these hardware limitations and utilizes non-convex optimization and deep learning methods that speed up and improve the accuracy of hologram synthesis from object wave reconstruction.

\section{Hologram Optimization}
There are two categories of 3D CGH optimization algorithms: superposition and iterative projection.  Superposition algorithms \enquote{decompose the desired 3D intensity into discrete 2D planes (segments) and digitally back-propagate each to the SLM plane. Each plane is considered independently, and the solution is the average of the phases of the resulting complex fields \cite{Zhang2:2017}.}  Iterative projection methods update the solution by projecting it onto a subspace computed from the set of constraints.  

Alternative projections (AP) can be achieved by a pair of elementary projections repeatedly occurring in the optimization, which construct an iterative computation loop. Specifically for CGH, alternating projections are applied to two enclosed sets associated with potential object solutions and potential hologram solutions.  The Gerchberg-Saxton (GS) algorithm, iterative Fourier-transform algorithm (IFTA), and iterative algorithm with angular spectrum (AS) theory are types of APs.  AS refers to a mathematical representation of a light waveform as a collection of plane waves with different spatial frequencies, essentially describing the distribution of light rays at various angles that make up the wavefront, which is crucial for calculating and reconstructing holographic images using techniques like the angular spectrum method.

\begin{figure}[H]
	\centering
	\begin{center} 
	\includegraphics[width=1\linewidth]{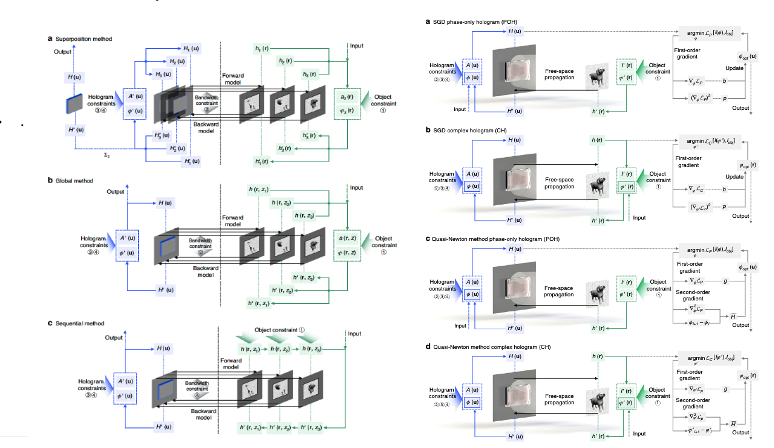}
	%\small \caption{} 	
	\caption{CGH Algorithms: Right side: (a) SGD-POH; (b) SGD-CH; (C) QN-POH; (d) QN-CH; Left side: (a) Superposition method; (b) Global Method; (c) Sequential method. \cite{Sui:2024}} 
	\label{fig:im7}  
	\end{center}  
\end{figure}

The inverse problem of hologram synthesis in CGH can also be cast as the optimization of a parameterized objective function requiring minimization with respect to its parameters. Since the choice of the objective function is often stochastic and differentiable with respect to its parameters, stochastic gradient descent (SGD) is considered as an efficient and effective first-order gradient descent framework for optimization. For first-order gradient descent, there is stochastic gradient descent (SGD) with single Fourier-transform propagation and SGD with angular spectrum theory. Second-order gradient descent can be implemented by the quasi-Newton (QN) method with single Fourier-transform propagation, which minimizes the loss function by constructing and storing a series of matrices that approximate the Hessian or inverse Hessian matrix of the loss function.  These algorithms are illustrated in Figure \ref{fig:im7}

\subsection{Fourier Transform Holography}

\ \ \ Holography solves the problem of recovering the phase distribution by adding a reference wave with a known distribution with the unknown object wave \cite{Mustafi:2023}.  In conventional holography, a reference wave is added to a complex-valued object wave in real space.  The reference wave is created by the diffraction from a reference object.  In Fourier transform holography (FTH), a reference wave is added to the complex-valued spectrum of any object wave in Fourier space.    

Both methods allow the complete distribution of a complex-valued object signal to be captured.  The object wave originates from a point-like source (e.g. a small aperture or a point-like scatterer) in real-space as does the reference source which could be a tiny particle that is present in the object plane.  The wave diffracted by the object interferes with the reference wave in the far field, and the resulting interference pattern generates the Fourier transform hologram.  

The \enquote{contrast of the resulting interference pattern (the hologram) is proportional to the difference between the phases of the object and the reference waves.  Thus, by knowing the phase distribution of the reference wave, it is possible for the phase distribution of the object distribution, to be reconstructed from the hologram \cite{Mustafi:2023}.} The object reconstruction is generated by computing the inverse Fourier transform of the recorded hologram.    

Optimization algorithms are designed to retrieve a CGH $H(\boldsymbol{u}) = A(\boldsymbol{u})e^{j \phi(\boldsymbol{u})}$ from a specific object wave $h(\boldsymbol{u}) = a(\boldsymbol{u})e^{j \psi(\boldsymbol{u})}$ in an inverse problem solving CGH situation, where $a(\boldsymbol{r})$ is the amplitude of the object waveform, $\psi(\boldsymbol{r})$ is the phase, $\boldsymbol{r} = (x,y)$ and $\boldsymbol{u} = (u,v)$ are the position vectors on the object plane and the hologram plane, respectively.  The optimization is restricted by several constraints: 1. object intensity $a^2(\boldsymbol{r}) = I_{obj}(\boldsymbol{r})$; 2. finite bandwidth $\Delta_{H} < \infty$ and $\Delta_{h} < \infty$; 3. finite spatial scale $L_{u} < \infty$ and $L_{v} < \infty$; and hologram intensity for POHs: $A^2(\boldsymbol{u}) = I_{hol}(\boldsymbol{u})$ 
    A POH can be synthesized by constructing an objective function $\mathcal{L_{P}}$
and solving the minimization problem:
\begin{equation}
    \phi_{opt} = \underset{\phi}{\text{argmin}} \ \mathcal{L_P}[I(\phi),I_{obj}]
\end{equation}
where $\mathcal{L_P}$ is the loss function that computes the difference between the reconstructed intensity $I(\phi)$ and the object intensity $I_{obj}$.  $I(\phi)$ denotes the reconstructed intensity as a function with respect to the POH $\phi$.   Since the reconstructed intensity $I_{\phi}$ and reconstructed wave $h(\phi)$ are connected by the ill-conditioned relationship $I(\phi) = \lvert h(\phi) \rvert^2$, the optimization synthesis problem is non-convex.  In the SGD method, the loss can be minimized with the Adam algorithm where the algorithm updates the exponential moving averages of the partial derivative $\nabla_{\phi} \mathcal{L_P}$.

For the synthesis of CH using SGD, the calculated gradient must be a real value.  Sui et. al. \cite{Sui:2024} resolve this issue by switching the output onto the object plane instead of the hologram plane since the object phase is the only floating parameter of the optimization model.  The minimization problem is:
\begin{equation}
    \psi_{opt} = \underset{\phi}{\text{argmin}} \ \mathcal{L_C}[I(\psi'),I_{obj}]
\end{equation}
where $\mathcal{L_C}$ is the loss function, and $I(\psi')$ is the reconstructed intensity function with respect to the object phase $\psi'$.

\subsection{Coherent Diffraction Imaging}
\ \ \ In coherent imaging, information about the phase of the propagating light wave sample that reaches the detector is lost.  The \enquote{sample} refers to the combined arrangement of the object wave with the reference wave.  The detector can only record wave intensity, namely, the rate at which the wave transfers energy over which the energy is spread (the ratio of power to unit area.)  The missing phase distribution is essential because it contains information about the scattering events that have taken place inside the sample \cite{Mustafi:2023}.   However, coherent diffraction imaging (CDI) is another lenless solution to the phase problem. 

In CDI, diffracted (or scattered) waves are recorded in the far field which allows \enquote{for the acquisition of high-resolution information \cite{Mustafi:2023}.} The resulting diffraction wavefront pattern is invariant to the lateral shifts of the sample and is given by the Huygens-Fresnel integral transform.  As result, the high-order diffraction signal is preserved.  Numerical iterative methods are often used to recover the missing phases.   

3D display can be realized using 3D imaging technologies such as volumetric 3D display, light field display, binocular vision display, structured light imaging, and time-of-flight imaging.  However, since holograms contain both the amplitude and phase information (which can completely reconstruct the intensity and depth) of the 3D scene, 3D holographic display is considered the ultimate solution for 3D display \cite{Song:2024}.  However, during optical holographic reconstruction, reducing the coherence of the light source is necessary since coherent light sources are one of the main causes of speckled noise generation. White-light recording has found to be the best optical solution for speckled noise \cite{Bianco:2018}.
Curtis et. al.

\begin{figure}[h!]
	\centering
	\begin{center} 
	\includegraphics[width=1\linewidth]{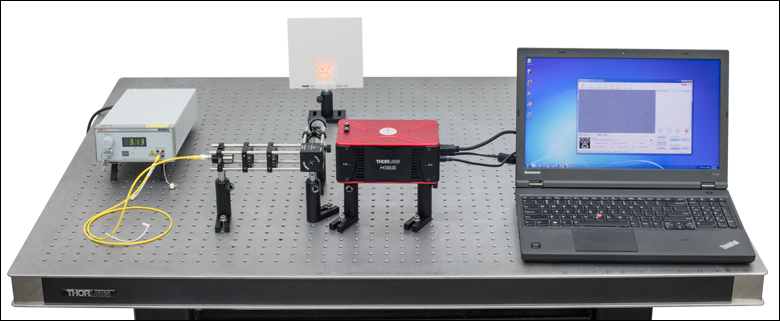}
	%\small \caption{} 	
	\caption{Physical SLM setup for CGH \cite{Thorlabs:2024}} 
	\label{fig:slm2}  
	\end{center}  
\end{figure}

Yu et. al. \cite{Yu:2023} introduce 3D scattering-assisted dynamic holography (3D-SDH) that enables ultra high density multi-plane projection and depth resolution.  The method overcomes two long-existing problems in current DH methods that limit 3D quality: low axial resolution and high interplane crosstalk.  3D-SDH combines an SLM with a diffuser that enables multiple image planes to be separated by a much smaller amount without being constrained by properties of the SLM.  SLMS spatially module a coherent beam of light. They can either be reflective or transmissive depending on the design purpose. There are two SLM types: optically addressed (convert incoherent light to spatial modulation) and electrically addressed (convert electrical signals to spatial modulation.).   SLMs include liquid crystals, deformable surfaces, and digital micromirror devices (DMD).  
\begin{figure}[h!]
	\centering
	\begin{center} 
	\includegraphics[width=1\linewidth]{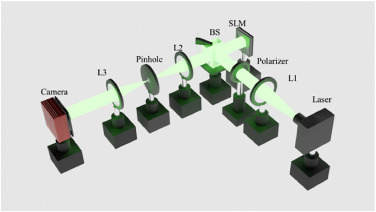}
	%\small \caption{} 	
	\caption{Optical display components \cite{Shen:2024}} 
	\label{fig:slm}  
	\end{center}  
\end{figure} 
 Figure \ref{fig:slm} illustrates the optical display system for CGH and Figure \ref{fig:slm2} shows actual physical set-up.
 In the CGH imaging system, a coherent laser beam light source emits a beam of light.  The light beam is split into a reference beam and an an object beam with a beam splitter.  The object beam illuminates the sample, causing light waves to diffract and scatter.  The object beam and the reference beam are recombined (superimposed) at a digital sensor (like a CCD camera) to create an interference pattern which is recorded  as a digital hologram.  The digital hologram is processed by a computer using various optimization or deep learning algorithms to calculate the phase and/or amplitude information of the light waves and reconstruct a 3D image of the object.  The hologram is then displayed on a spatial light modulator (SLM) allowing the 3D image to be reproduced and projected in real space.

%\begin{figure}[H]
%	\centering
%	\begin{center} 
%	\includegraphics[width=1\linewidth]{slm.png}
%	%\small \caption{} 	
%	\caption{Source: \cite{}} 
%	\label{fig:slm}  
%	\end{center}  
%\end{figure} 

Various researchers have applied deep learning methods to computational holography \cite{Chen:2021, Liu:2021, Zeng:2021, Khan:2021, Shimobaba:2022} accelerating the growth in computational holography as graphical processing units (GPUs) have increased in speed and power.  Deep learning methods have been shown to outperform previously physically-based calculations using lightwave simulations and signal processing. 

Song, et. al. \cite{Song:2024} propose a real-time 3D computer-generated holography (CGH) display system of the real-world environment that operates at 22 frames per second. The CGH system receives input intensity and depth maps of 3D scene/object information (acquired from an RGB-D camera and depth camera, respectively)  into a convolutional neural network (CNN). The 3D scene/object is divided into $J = 30$ layers sing the layer-based method with an exit focal length of $z = 0.2 m$.  The trained network generates a reconstructed CGH with a resolution of 1024 $\times$ 1024 in 14.5 ms. 

In general, 3D CGH is implemented by encoding the phase of a coherent wavefront in the Fourier space (pupil plane) of an imaging system with a SLM. \cite{Zhang:2017}
Computing the 2D Fourier phase mask that generates a specific 3D intensity is usually an ill-posed or non-convex optimization problem with non-unique solutions in a high-dimensional space (millions of pixels) that cannot be fully explored. Most 3D intensity patterns are limited to the coherent diffraction limit set by the numerical aperture (NA), the limited degrees of freedom (DOF) for patterning, and the fact that intensity must be conserved at each depth plane \cite{Zhang:2017}.  Furthermore, there are hardware limitations such as the SLM dynamic range, pixel pitch, and pixel count.  Thus, 3D CGH solutions are approximate where there is no exact solution. 

\begin{figure}[h]
	\centering
	\begin{center} 
	\includegraphics[width=1\linewidth]{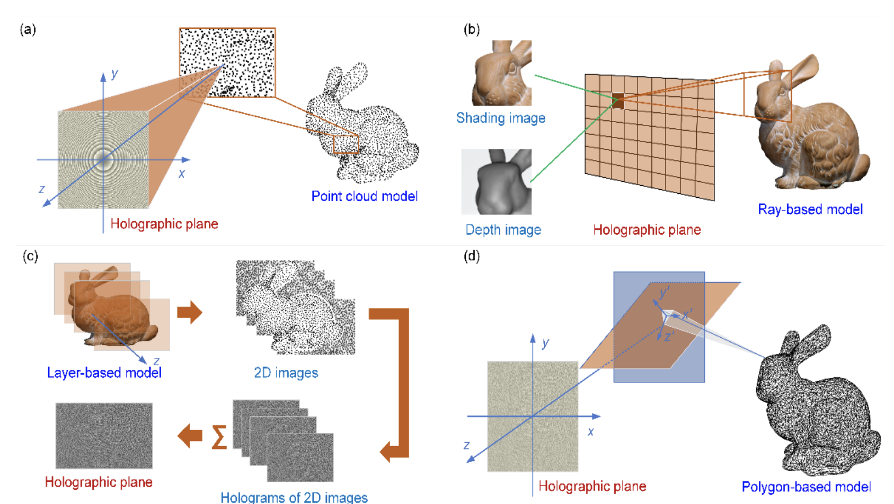}
	%\small \caption{} 	
	\caption{Hologram synthesis from different primitives. (a) point-based CGH method; (b) ray-based CGH method; (c) layer-based CGH method; and (d) polygon-based CGH method. \cite{Cao:2022}} 
	\label{fig:prim}  
	\end{center}  
\end{figure}

There are various types of CGH including point-based methods, polygon-based methods, layer-based methods, ray-tracing, geometric primitives, basis functions, and holographic stereograms as illustrated in Figure \ref{fig:prim}.  
There are various phase optimization methods: (1) iterative, (2) non-iterative, (3) direct computation, (4) combined iteration and non-iteration, (5) Wirtinger flow, and (6) deep-learning.  Iterative methods include the Gerchberg-Saxton (GS) algorithm, Fienup algorithm, and error diffusion algorithm.  Non-iterative methods include random phase, sampling methods (e.g. sampled-phase-only hologram, complementary phase-only hologram, and adaptive down-sampling masks), patterned phase-only hologram and quadratic phase, double-phase method (complex amplitude modulation) \cite{Hsueh:1978}, and non-random phase-free method.  There are also Fourier methods such as propagation fields that include the Fresnel diffraction method and angular spectrum method (ASM).

Iterative algorithms typically start with an approximation of the target hologram and continuously optimize the approximate hologram through a series of repeated operations until the reconstructed image obtained by the approximation meets certain error requirements \cite{Pi:2022}.  The error diffusion algorithm \enquote{iterates between the pixels of the hologram plane in turn, rather than between the hologram plane and the object image plane, without any information of the object image, only the complex amplitude hologram itself can be directly operated on it and a pure phase hologram can be calculated \cite{Jack:2022}.} 

There are two categories of 3D CGH optimization algorithms: superposition and iterative projection.   Superposition algorithms \enquote{decompose the desired 3D intensity into discrete 2D planes (segments) and digitally back-propagate each to the SLM plane. Each plane is considered independently, and the solution is the average of the phases of the resulting complex fields \cite{Zhang2:2017}.}  Iterative projection methods update the solution by projecting it onto a subspace computed from the set of constraints.   

The Gerchberg-Saxton (GS) is the most widely-used iterative CGH algorithm and has two iterative versions: sequential and global (corresponding to batched SGD).  GS retrieves phase information from the measured intensities on two related planes (the source and target planes).  Other iterative methods include stochastic gradient descent (SGD) and quasi-Newton (QN).  GS is an iterative error reduction method that simulates the propagation of a coherent wave back and forth between the complex field image plane and SLM. These algorithms work well for single-plane intensity patterns, but diverge when attempting to generate physically infeasible patterns.  To address this issue, Zhang, et. al. \cite{Zhang2:2017} introduce the Non-Convex Optimization for VOlumetric CGH (NOVO-CGH) algorithm that combines a customizable cost function with an optimization strategy inspired by prior work on phase retrieval.

Curtis, et. al. introduce Dynamic CGH (DCGH), a light sculpting method that modulates light both spatially and temporally to average out \enquote{uncontrolled interference of constructive and destructive waves} that can cause speckled noise \cite{Curtis:2021}. DCGH generates realistic transformative 3D images that cannot be achieved with single-frame CGH.  DCGH uses a DMD and renders speckle-free 3D images by rapidly displaying a superposition of jointly optimized coherent waves.  A collimated  650 nm laser source illuminates the effective area of a DMD, a computer controlled 2D array of micro-mirrors that can be individually addressed and flipped on demand between on-off positions.   %The DMD applies a custom binary  

\subsection{Volumetric Reconstruction}
To generate a high-quality hologram, one must reconstruct a 3D volumetric object that is robust to noise, outliers, misalignment, and missing data.  In this project, we propose a pipeline methodology and implementation to reconstruct a high-quality 3D volumetric object from a point cloud, and then generate a high-quality 3D CGH hologram that is robust to these artifacts. For purposes of this project, we will focus on MRI scans. 

Point clouds are obtained from range scanning devices that scan a physical object into their point coordinates in free space.  Scanning generates multiple range images that each contain 3D points for different pats of the model in local coordinates of the scanner. The registered to the point cloud. There are many approaches for 3D reconstruction of an object.  Multi-stereo view is one approach in which given several images of an object taken from a range scanner or camera, one computes a representation of its 3D shape using depth maps, meshes, point clouds, patch clouds, volumetric models, etc.    

Marching cubes is a popular 3D reconstruction surface rendering algorithm for medical imaging \cite{Lorensen:1987}.  It creates triangular models of constant density surfaces from 3D medical data which is processed in a scan-line order.   Triangular vertices are calculated using linear interpolation.  Volume data values are recorded as voxels on a discrete 3D grid.  A voxel is a volume element representing a value on a regular grid in 3D space.  Each voxel has 8 points or vertices.  Voxels do not have their coordinates explicitly encoded with their values.  Rather, rendering systems infer a voxel's coordinates based on its position relative to other voxels \cite{Sakamoto2:2017}.

The marching cubes algorithm determines how the surface (corresponding to a user-specified value) intersects the cube and creates triangles.  The normals to the surface at each vertex of each triangle is computed.  To determine the surface intersection in a cube, one assigns a 1 to a cube's vertex if the data value at that vertex exceeds or equals the value of the surface being constructed.   These vertices are inside (or on) the surface.  Cube vertices with values below the surface have a value of 0 and are outside the surface.  The surface intersects those cube edges where one vertex is outside the surface (1) and the other is inside (0).  Due to two different symmetries of the cube, the algorithm triangulates $2^8 = 256$ cases into 14 patterns \cite{Lorensen:1987}.   

\begin{figure}[h!]
	\centering
	\begin{center} 
	\includegraphics[width=0.5\linewidth]{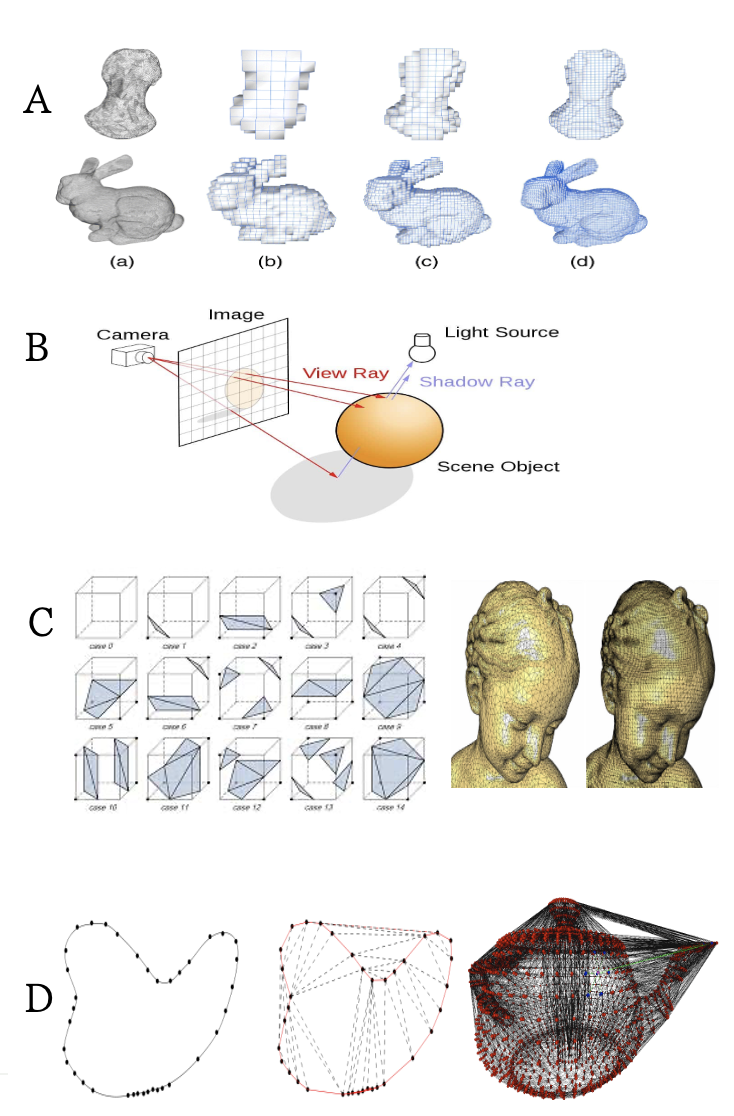}
	%\small \caption{} 	
	\caption{Volumetric Reconstruction Methods. (A) mesh voxelization; (B) ray tracing; (C) marching cubes; (D) Delaunay triangulization} 
	\label{fig:recon4}  
	\end{center}  
\end{figure}

A volume can be visualized either by direct volume rendering or by extracting polygon iso-surfaces that follow the contours \cite{Sakamoto2:2017}.   Direct volume rendering requires every voxel value to be mapped to an opacity and a color.  Ray tracing and ray casting can be used to render the volume directly.   The marching cubes algorithm can be used for iso-surface extraction.  Figure \ref{fig:recon4} illustrates different volumetric reconstruction methods.

\begin{figure}[h!]
	\centering
	\begin{center} 
	\includegraphics[width=0.9\linewidth]{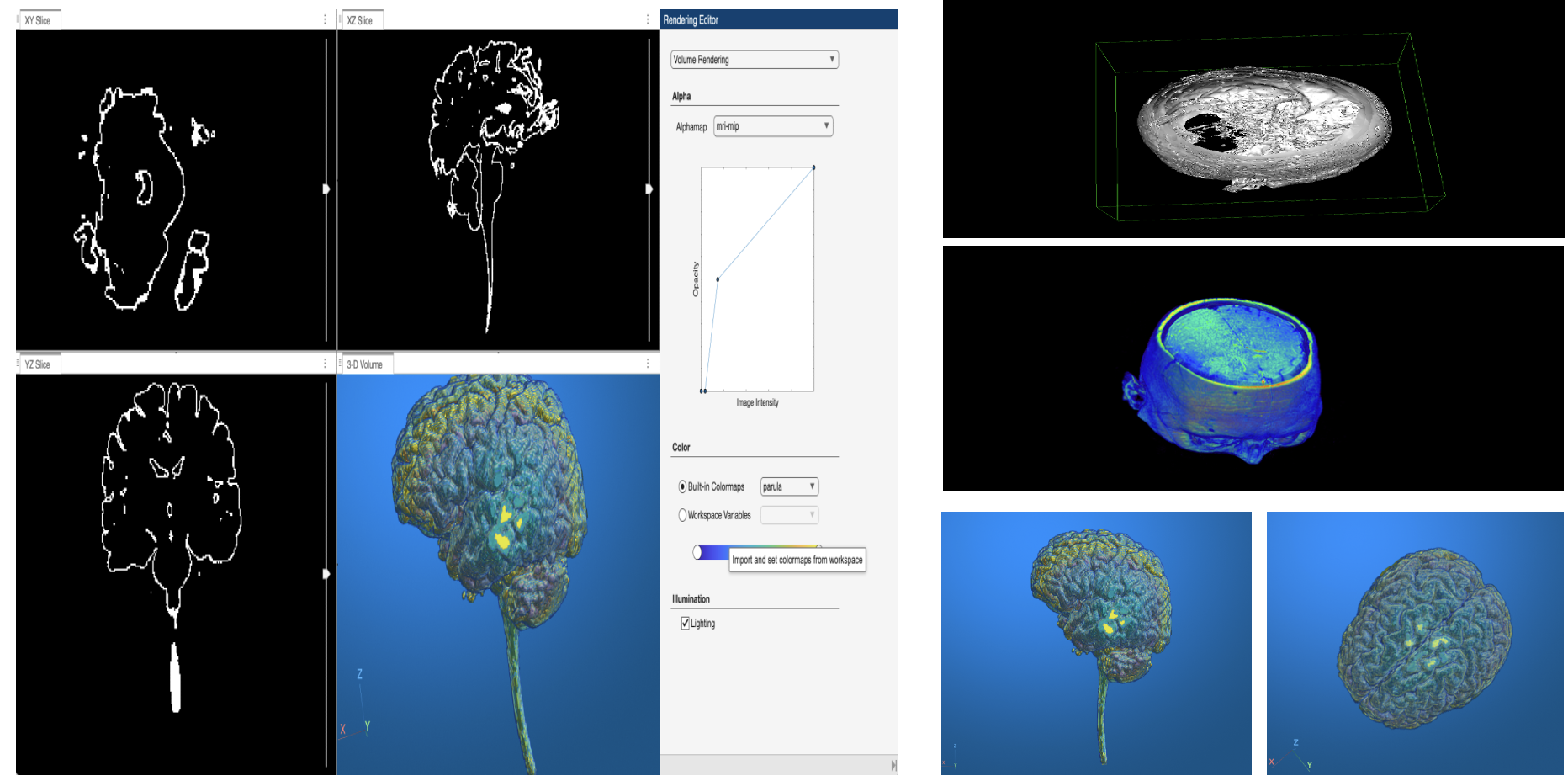}
	%\small \caption{} 	
	\caption{Volumetric rendering in Matlab. Left: volumetric brain displayed along different axes. Right: (a) top: volume generated from MRI DICOM brain slices. (b) volume display in skull; (c) volume rotation to view top of brain.} 
	\label{fig:vol3}  
	\end{center}  
\end{figure}

Once the volumetric image is rendered, a method for generating its hologram display is required.   Two such methods are the polygon-based method and maximum intensity projection based (MIP-based) method \cite{Sakamoto2:2017}.  In the polygon-based approach, the marching cubes algorithm is used to generate polygon models of the volume data. For MRI digital imaging and communication in medicine (DICOM) images, the DICOM viewer OsiriX generates the marching cube polygons which can be exported \cite{Sakamoto2:2017}. To calculate the CGH with polygon data, ray-tracing method can be used where ray-tracing determines the intersections between rays cast from the view point and triangle patches of the polygon model \cite{Sakamoto2:2017}.  Figure \ref{fig:vol3} illustrates 3D volume rendering of point cloud brain data in Matlab using ray-tracing.
The MIP projects in the visualization plane with voxels with maximum voxel value that fall in the way of parallel rays traced from the viewpoint to the plane of projection. \cite{Sakamoto2:2017}  Lu and Sakamoto \cite{Sakamoto2:2017} reconstruct the MIP data with depth coordinates on the basis of the positions relative to  other voxels to improve the sense of depth.   With this method, only one MIP result can be obtained in each direction and the MIP result is a single color image.   There are three parts to their volume rendering method.  First, there is light propagation from the light source to voxel $i$, $i = 1,\dots,N.$.  This can be represented by the attenuation equation:
\begin{equation}
    I_{d_{i}} = I_{d_{0}}\Pi_{j=0}^{i}(1 - \alpha_{j})
\end{equation}
where $I_{d_{0}}$ is the intensity of the light source, $\alpha_{j}$ is the opacity value of voxel ${j}$, and $I_{d_{i}}$ is the incident light of the second step.  The intensity of reflected light can be calculated using the Lambertian model:
\begin{align}
    I_{r_{i}} &= k_{d_{i}}I_{d_{i}}max(0,\boldsymbol{L}\cdot \boldsymbol{N}) \\
              &= k_{d_{i}}I_{d_{i}}\text{cos}(\theta) 
\end{align}
where $k_{d_{i}}$ corresponds to the ratio of Lambert light, the vector $\boldsymbol{L}$ is the unit vector pointing to the light source, vector $\boldsymbol{N}$ is the normal unit vector (its values are the gradient values of $\alpha$ in the x,y, and z direction, and $\theta$ is the angle between $\boldsymbol{L}$ and $\boldsymbol{N}$.  

Third, is the light propagation from voxel $i$ to the viewpoint and then calculating the hologram for the rendered volume data.  One can apply the point light source to generate hologram data.  Since the intensity of reflected light does not depend on viewpoint direction, we can replace $d_{i}$ with the final reflected light intensity of voxel $i$, $r_{L_{i}}.$  For point light source $i$, the complex amplitude distributions $\mu(\zeta,\eta)$ are computed by:
\begin{equation}
    \mu(\zeta,\eta) = \sum_{i=1}^{N} \mu_{i}(\zeta,\eta) 
\end{equation}
where
\begin{align}
    \mu_{i}(\zeta,\eta) = \frac{I_{r_{L_{i}}}}{r_{i}}e^{-j(2\pi /\lambda + \phi_{i})} 
\end{align}
$\phi_{i}$ is the random phase of point light source $i$, and $d_{i}$ is the distance between the point light source $i$ and the pixel on the hologram plane. $\lambda$ is the wave length of light.  
Denote the visual object point $\boldsymbol{r_{i}} = (x_{i},y_{i},z_{i})$ and the coordinates of the point light source $\boldsymbol{r}'_{i} =(x'_{i},y'_{i},z'_{i})$.
The light waves on the hologram plane for the 
\begin{equation}    
    r_{i} = \sqrt{(x - x'_{i})^2 + (y - y'_{i})^2 + (z'_{i})^{2}}
\end{equation}
$I^{n}_{r_{L_{i}}}$ is the amplitude of point light source $i$ for viewpoint $n$, which represents (R,B,G) information \cite{Lu2:2019}.  The intensity of the inference pattern is calculated as:
\begin{equation}
    I(\zeta,\eta) = \lvert \mu(\zeta,\eta) + R(\zeta,\eta) \rvert^{2}
\end{equation}
where $R(\zeta,\eta)$ is the parallel reference light.   In 3D hologram generation, the impulse response $h(x,y;z)$ in a paraxial geometry can be assumed to be that for Fresnel diffraction given in Eq. \ref{eq:fresnel}.
%3D image reconstruction form a 

%where $I_{d_{0}}$ is the intensity of the light source, $\alpha_{j}$ is the opacity %value of voxel ${j}$, and $I_{d_{i}}$ is the incident light of voxel $i$. 
%Unlike the polygon-based method, the MIP-based holographic display method is a direct volume 

\section{Fourier Transform Optical System}

%\section{Fresnel transform}
In Fourier transform holography (FTH), the exit wave behind the sample $u(x,y)$ is given by the transmission function of the sample.   The transmission function characterizes the interaction between the incident wave and the sample wave and it is assigned to a plane:
\begin{equation}
t(x,y) = \text{exp}[(-a(x,y) + i\varphi(x,y)]
\end{equation}
where $\text{exp}[-a(x,y)]$ is the amplitude of the complex-valued transmission function, $a(x,y)$ is the function that describes the absorption properties of the sample, $\varphi(x,y)$ is the function that describes the phase added by the sample into the passing wave, and $(x,y)$ is the coordinate in the sample plane \cite{Latychevskaia:2019}. Thus, $u(x,y) = u_{o}(x,y)t(x,y)$ and can be represented as:
\begin{equation}
    u(x,y) = u_{o}(x,y) + u_{r}(x,y)
\end{equation}
where $u_{o}(x,y)$ is the object wave term and $u_{r}(x,y)$ is the reference wave term where the subscript $r$ is with reference to the sample plane. The transmission function in the object plane can be written as 
\begin{align}
    t(x,y) &= 1 + \frac{u_{r}(x,y)}{u_{o}(x,y)} \\
           &= 1 + o(x,y) 
\end{align}
where 1 corresponds to the transmittance in the absence of the object, and $o(x,y) = u_{r}(x,y)/u_{o}(x,y)$ is a complex-valued function that describes the perturbation caused by the presence of the object wave.
%and $\boldsymbol{r} = (x,y,z=0)$ is the coordinate in the sample plane.   
%Though macroscopic 3D sceneries generate spatially incoherent optical fields, most configurations rely on lasers.  Due to the improvement of the spatial resolution of charged-coupled device (CCD)    Optical correlation Mmthods and algorithms have been useful in the design of 2D pattern-recognition applications \cite{Javidi:2000}.  Various methods have been proposed to extend optical correlation methods to 3D object recognition. 
    We can simulate the hologram using a Fresnel transform.  Denote the incident wave $U_{in}(x,y)$ in the object plane as
\begin{equation}
    U_{in}(x,y) = \frac{\text{exp}(ik\boldsymbol{r})}{\boldsymbol{r}} 
\end{equation}
\normalsize
where $\boldsymbol{r} = (x,y,z)$ is a vector pointing from the source point to a point in the object, and $z$ is the distance between the source and object plane.  The exit wave $U_{exit}(x,y)$ behind the object is given by the product of the incident wave and the transmission function $t(x,y)$:
\small
\begin{equation}
    U_{exit}(x,y) = U_{in}(x,y)t(x,y) = \frac{\text{exp}(ik\boldsymbol{r})}{\boldsymbol{r}}t(x,y)
\end{equation}
\normalsize
    The Fresnel-Kirchhoff diffraction formula characterizes the propagation of the wave toward the detector:
\small
\begin{align}
    &U_{detector}(X,Y) = \\ &-\frac{i}{\lambda} \int \int \frac{\text{exp}(ik\boldsymbol{r})}{\boldsymbol{r}}t(x,y) 
    \frac{\text{exp}(ik \lvert \boldsymbol{r} - \boldsymbol{R} \rvert)}{\lvert \boldsymbol{r} - \boldsymbol{R} \rvert}dxdy
\end{align} 
\normalsize
or written in Fourier or k-coordinates using the the Huygens-Fresnel integral transform:
\small 
\begin{align}
    U(\boldsymbol{R}) &= -\frac{i}{\lambda} \int \int u(\boldsymbol{r})\frac{\text{exp}(ik \lvert \boldsymbol{r} - \boldsymbol{R} \rvert)}{\lvert \boldsymbol{r} - \boldsymbol{R} \rvert}dxdy \\
     &\approx -\frac{i}{\lambda R} \int u(\boldsymbol{r})\text{exp}\bigg (ik(R -\frac{\boldsymbol{K}\boldsymbol{r}}{k})\bigg)d\boldsymbol{r} 
    \label{eq:fresnel}
\end{align} 
\normalsize
where $\boldsymbol{R} = (X,Y,Z)$ is a vector point from the source to a point on the the detector (the coordinate in the far field),  $k = \frac{2 \pi}{\lambda}$ is the wavenumber, $\lambda$ is the wavelength, $\lvert \boldsymbol{r} - \boldsymbol{R} \rvert$  is the distance between a point in the object plane and a point in the detector plane, and
\begin{equation}
    \boldsymbol{K} = k\frac{\boldsymbol{R}}{R} = \frac{2 \pi}{\lambda}\bigg (\frac{X}{R},\frac{Y}{R},\frac{Z}{R} \bigg ) 
\end{equation} 
given $R = \lvert \rvert \boldsymbol{R} \lvert \rvert = \sqrt{X^2 + Y^2 + Z^2}$.
The integration in all integrals is performed over a finite area in the sample (or detector) plane.  Equation \ref{eq:fresnel} can be converted to Cartesian coordinates.  Using the Taylor series expansion:
\begin{equation}
    \lvert \boldsymbol{r} - \boldsymbol{R} \rvert \approx Z + \frac{(x-X)^2 + (y-Y)^2}{2Z}
\end{equation}
\normalsize
so that
\small
\begin{align}
    &U(X,Y) \approx -\frac{i}{\lambda Z}\text{exp}\bigg [\frac{2 \pi i}{\lambda}\bigg (Z + \frac{(X^2 + Y^2)}{Z}   \bigg  ) \bigg ] \cdot  \nonumber \\ 
    & \int \int u(x,y) \text{exp} \bigg (-\frac{2 \pi i}{\lambda Z}(xX + yY) \bigg )d\text{x}d\text{y}
\end{align}

\normalsize
where the approximation \small $\lvert \boldsymbol{r} - \boldsymbol{R} \rvert \approx R  - \frac{\boldsymbol{r \cdot R}}{R}$.   
\normalsize At \small $Z^2 >> X^2 + Y^2$, \normalsize the vector components can be approximated as $\boldsymbol{K} \approx \frac{2 \pi}{\lambda} \big(\frac{X}{Z},\frac{Y}{Z},1 \big )$, yielding
\small
\begin{align}
    &U(K_{x},K_{y}) \approx -\frac{i}{\lambda R}\text{exp}(ikR) \cdot \nonumber \\ 
    &\int u(x,y)\text{exp}[-i(xK_{x} + yK_{y})]d\text{x}d\text{y}
\end{align}
\normalsize
and 
\small
\begin{equation}
    I_{H}(K_{x},K_{y}) \approx \bigg \lvert \int \int u(x,y)\text{exp}[-i(xK_{x} + yK_{y}]dxdy \bigg \rvert^2 \nonumber  
\end{equation} 
\begin{equation}
= \lvert \mathcal{F}[u(x,y)] \rvert^{2}
\end{equation}
    \normalsize
    Thus, in Fourier or Cartesian coordinates, the intensity distribution in the far field is given by the squared amplitude of the the Fourier transform of the exit wave.  
\begin{figure*}[ht!]
	\centering
	\begin{center} 
	\includegraphics[width=0.8\linewidth]{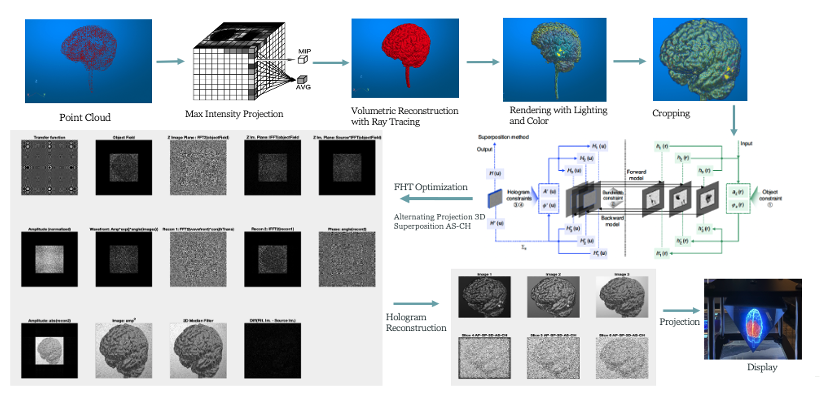}
	%\small \caption{} 	
	\caption{Proposed CGH Pipeline.  Top: (a) point cloud data as a mesh is input into (b) MPI algorithm that renders (c) volumetric object; (d) lighting is added; and (e) cropping; Middle: (f) cropped volumetric is input into the (g) AP method for optimization.  During optimization, the (h) algorithm parameters such as phase, transfer function and amplitude can be visualized in the pane;  Bottom: (i) The object wave (hologram + wavefront) is reconstructed as slices; The optimized hologram is projected onto (j) a 3D display.}
	\label{fig:slm}  
	\end{center}  
\end{figure*} 
The hologram intensity distribution $I_{H}$ is:
\small
\begin{align}
    &I_{H}(X,Y) \approx \nonumber \\ &\bigg \lvert \int \int u(x,y) \text{exp} \big [-\frac{2 \pi i}{\lambda Z}(xX + yY) \big ]dxdy \bigg \rvert^{2} \nonumber \\ &= \lvert \mathcal{F}[u(x,y)] \rvert^{2}
\end{align} 

\normalsize
where $\mathcal{F}$ is the Fourier transform (FT).   The hologram can be formed as the squared amplitude of the wavefront in detector plane.   The relationship between the diffraction pattern parameters and the reconstructed field of view (FOV) is given by:
\small
\begin{align}
    &u_{rec}(x,y) = \nonumber \\ &\int \int I_{H}(X,Y)\text{exp} \bigg [\frac{2\pi}{\lambda Z}(Xx + Yx) \bigg ]dXdY
    \label{eq:12}
\end{align}
\normalsize
    which can be rewritten as the 2D inverse fast Fourier transform (IFFT):
    \small
    \begin{equation}
        u_{rec}(m,n) = \sum_{p=0}^{M-1} \sum_{q=0}^{N-1}I_{H}(p,q)\text{exp}\bigg [ 2\pi i \bigg (\frac{pm}{M} + \frac{qn}{N} \bigg) \bigg ]
        \label{eq:13}
    \end{equation}
    \normalsize
    where the object coordinates $(x,y)$ are digitized as $x \rightarrow \Delta_{x}m$, $y \rightarrow \Delta_{y}n$, the images coordinates (X,Y) are digitized as $X \rightarrow \Delta_{X}p, Y \rightarrow \Delta_{Y}q$, $\Delta_{x} = \Delta_{y}$ is the pixel size in the sample plane, $\Delta_{X} = \Delta_{Y}$ is the pixel size in the detector plane, and $m,p = 0,\dots,M-1$,$n,q = 0,\dots,N-1$ are the pixel numbers.   From equations \ref{eq:12} and \ref{eq:13}, it follows:
    \begin{equation}
        \frac{1}{\lambda Z}\Delta_{X}\Delta_{x} = \frac{1}{N}
    \end{equation}
    and the reconstructed field of view:
    \begin{equation}
        s_{0} = N \Delta_{x} = \frac{\lambda Z}{\Delta_{X}}    
    \end{equation}
    
    The reconstructed field of view must be large enough to include four reconstructed distributions (the object's autocorrelation and its two centro-symmetric sidebands.)   Thus, $s_{0} = \frac{\lambda Z}{\Delta_{X}} > 4D$ where $D$ is the diameter (extent) of the object.  To prevent an overlap of the reconstructed-object and the object autocorrelation distributions, the object and reference must be separated by a distance $L > 1.5D$.
    Reconstruction of the object distribution can be reconstructed from its FT hologram by calculating the inverse FT (IFT) or FT of the hologram:
    \begin{equation}
        I_{H} = \lvert \mathcal{F}[u(x,y)] \rvert^{2}
    \end{equation}
    and 
    \begin{equation}
        IFT(I_{H}) = u(x,y) \circ u(x,y) 
    \end{equation}

\normalsize
where $\circ$ is the correlation operator.
The reconstruction of a digital hologram recorded with plane waves begin with illumination with the reference wave $u_{r}(X,Y) = \text{exp}(ik \boldsymbol{R})/\boldsymbol{R}$ followed by backward propagation to the object plane described by the inverse Fresnel-Kirchhoff diffraction formula:
\small
\begin{align}
    U(x,y) &\approx \frac{i}{\lambda} \int \int \frac{\text{exp}(ik\boldsymbol{R})}{\boldsymbol{R}}H(X,Y) \cdot \nonumber \\ &\frac{\text{exp}(-ik \lvert \boldsymbol{r} - \boldsymbol{R} \rvert)}{\lvert \boldsymbol{r} - \boldsymbol{R} \rvert}dXdY
\end{align}
\normalsize
where $H(X,Y) = \lvert U_{detector}(X,Y) \rvert^{2} = \lvert t(X,Y)   \ast h(X,Y) \rvert^{2}$ and $h(x,y)$ is the Fresnel function:
\begin{equation}
    h(x,y) = -\frac{i}{\lambda z}\text{exp}\big [\frac{i \pi}{\lambda z}(x^2 + y^2) \big ]
    \label{eq:fresenlf} 
\end{equation}

\section{Deep Learning CGH}

 Deep-learning CGH improves hologram reconstruction by rapidly eliminating twin-image and self-interference artifacts using only one hologram intensity.  In addition, deep learning CGH has following advantages over other optimization methods:
 1. Faster reconstruction:  Deep learning algorithms can reconstruct images from holograms much faster than traditional iterative methods, enabling real-time applications. 
 2. Improved image quality:  Deep learning can effectively remove noise and artifacts like twin images from reconstructed images, resulting in higher quality 3D visualizations. 
 3. Enhanced phase recovery: Deep learning models can accurately recover the phase information from a hologram, which is crucial for precise 3D reconstruction. 
Robustness to low-quality data: Deep learning can reconstruct good quality images even from noisy or low-quality holograms, which is particularly useful in challenging imaging conditions. 
4. Generalization to new data:   Trained deep learning models can be applied to reconstruct images from new samples that were not seen during training, making them versatile for different applications. 
Complex scene handling: deep learning can handle complex 3D scenes with multiple objects and occlusions more effectively than traditional methods.  However, deep learning methods have the disadvantage that they are computationally more expensive and resource intensive than standard optimization methods, require GPUs, and less interpretable.  
DeepNet \cite{Eybposh:2020}, HoloEncoder \cite{Wu:2021} and HoloNet \cite{Peng:2020} are convolutional neural networks (CNNs) for CGH that incorporate Fourier transform operations to extract frequency information from input data.  In DeepCGH, for instance, a trained CNN maps the target amplitude $A(x,y,z)$ to a feasible approximation of the desired illumination pattern in the image plane by estimating the complex field at z = 0, $\hat{P}(x,y,z=0)$. The complex field is the virtually propagated back to the SLM plane with an inverse 2D Fourier transform to yield the solution to the CGH problem, namely, SLM phase mask $\phi_{SLM}.$ HoloNet combines camera-in-the-loop holography with deep learning to correct for degradations of reproduced images of holographic displays due to aberrations of optical components, misalignment of beam spliters and lenses, uneven light distribution of a light source on the SLM, and quantized and non-linear light modulation of SLM.  \cite{Shimobaba:2022}.  %Figure \ref{fig:holo} illustrates HoloNet training.

\begin{figure*}[h!]
	\centering
	\begin{center} 
	\includegraphics[width=0.8\linewidth]{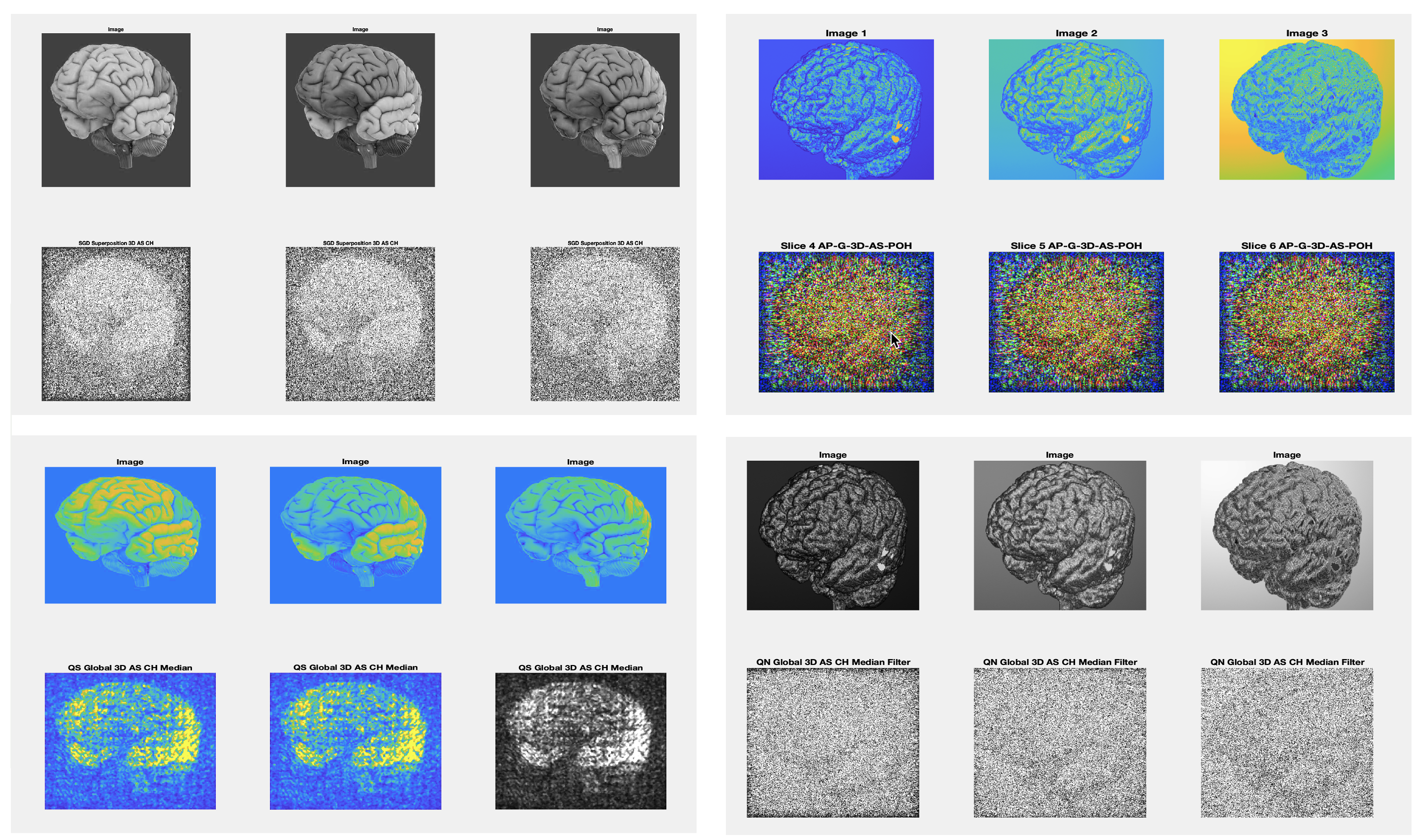}
	%\small \caption{} 	
	\caption{Optimized Reconstructed Brain Holograms.  (a) Upper left panel: 3D brain model image and optimization slices using SGD-SP-3D-AS-CH method; (b) Upper right panel: Volumetric brain image and optimization slices using AP-G-3D-AS-POH method; (c) Lower left panel: 3D brain model and image optimization slices using QN-G-3D-AS-CH method with median filtering;  (d) Lower right panel: volumetric brain image and optimization slices using QN-G-3D-AS-CH method with median filtering.} 
	\label{fig:im1}  
	\end{center}  
\end{figure*} 
\section{Methodology}
    \normalsize
    Point cloud data can be generated from 3D laser scanners and LiDAR (Light Detection and Ranging) sensors. To reconstruct a 3D volumetric object from a point cloud, we use maximum intensity projection (MIR) or marching cubes.  However, in Matlab, the \code{volshow} method uses ray tracing to render 3D volume data allowing for the visualization of volumetric images by calculating the color (pixel intensity) along a line of sight through the volume, taking into account factors like opacity and transfer functions set by the user.   
 \begin{figure}[h!]
	\centering
	\begin{center} 
	\includegraphics[width=1\linewidth]{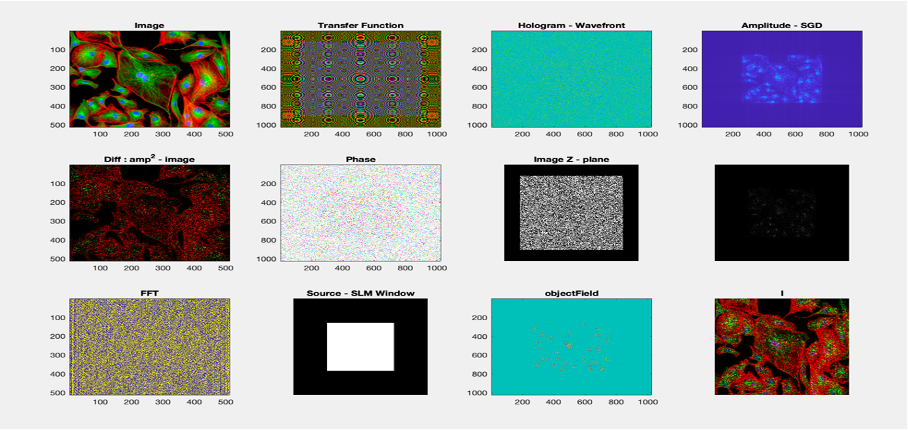}
	%\small \caption{} 	
	\caption{Fourier optic parameters during SGD SP-AS-CH method optimization. Top row (left to right): (a) original image; (b) transfer function; (c) hologram-wavefront; (d) squared amplitude; Middle row: (e) squared amplitude-image; (f) phase;  (g) image z-plane; (h)  Bottom row: (i) FFT  (j) PSF (SLM window); (j) objectField; (h) recon. object wave (hologram + wavefront).}
	\label{fig:panel2}  
	\end{center}  
\end{figure} 
    To generate a CGH, we use Fresnel transforms using both non-convex optimization \cite{Sui:2024} and deep learning.  In particular, we use alternative projection, stochastic gradient descent (SGD), and quasi-Netwton algorithms for both phase-only holograms (POH) and complex-holograms (CH).  We show that these algorithms are quite powerful for reconstructing 3D holograms from volumetric data and evaluate their reconstruction performance by measuring the mean squared error (MSE), root means squared error, and peak signal-to-noise ratio (PSNR) both with and without median filtering.  We also compare these results using the reconstructed hologram using HoloNet. The method that generates the highest PSNR is chosen for the proposed pipeline.

\begin{figure}[h!]
	\centering
	\begin{center} 
	\includegraphics[width=0.9\linewidth]{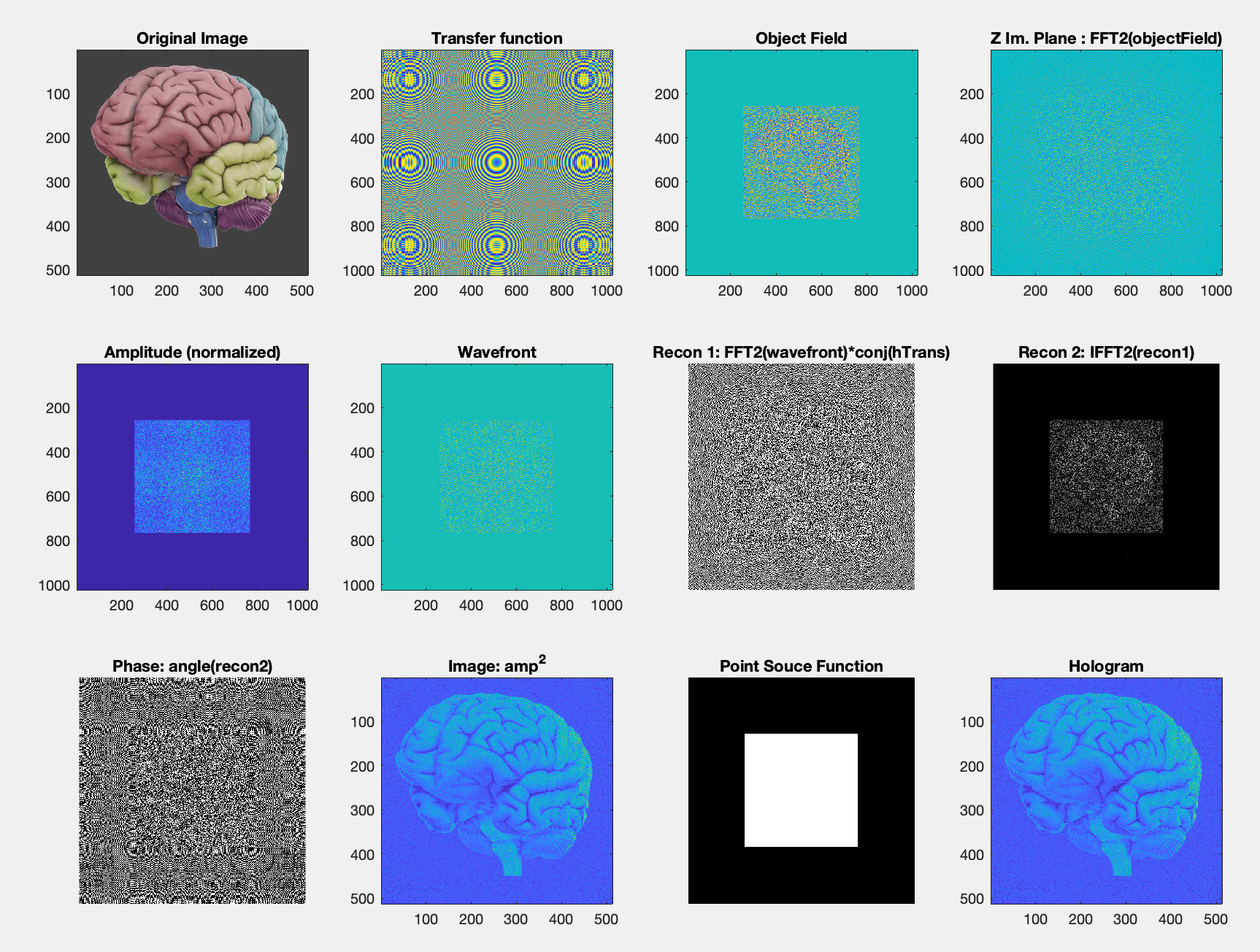}
	%\small \caption{} 	
    \caption{Fourier optic parameters during AP SP-AS-CH method optimization. Top row (left to right): (a) original image; (b) transfer function; (c) object field; (d) FFT(objectfield); Middle row: (e) amplitude (normalized); (f) wavefront;  (g) 1st reconstruction; (h) 2nd reconstruction.  Bottom row: (i) phase  (j) squared amplitude (k) PSF (SLM window) (l) recon. object wave (hologram + wavefront).}
	%\caption{Optimized Holograms. Left: (a) top row: volumetric brain slices. (b) optimized reconstruction slices. Right: (a) top row: 3D brain model image slices. (b) bottom row: optimized reconstruction slices.} 
	\label{fig:im2}  
	\end{center}  
\end{figure}

    We test the methodology using MRI point cloud data of a brain, MRI brain slices, and an image of cell taken using fluorescent microscopy.   The complete methodology is given as follows:
    1. Acquire point cloud and MRI brain slice data (.ply and DICOM files)
    2.  Read and import this scanned data into Matlab
    3. Convert 2D images into .mat files
    4. Perform 3D volumetric reconstruction 
    5. Set CGH parameters for the simulated spatial light modulator resolution dimensions, pixel pitch, and wavelength. 
    6. Input the 3D volumetric object into Fourier transform CGH optimization algorithms
    7. Run optimization and reconstruction algorithms with and without 3D median filtering.
    8.  Calculate depth, MSE, RMSE, and PSNR
    9.  Save reconstruction image and generate .gif images
    10. Run HoloNet deep learning training, perform prediction on unseen images used and calculate MSE loss
    11. Compare performance (results) of algorithms and determine best CGH pipeline/methodology.
    12. Project and display 3D hologram using Pepper's ghost effect.

\section{Simulation Experiment}

%\subsection{}
We assume that we are using a spatial light modulator with a pixel pitch of $3.74e-3$ \ mm, a focal distance of $f = 3.74e-3$ \ mm, and spatial window size resolution of $N_{x} \times N_{y} = 512 \times 512$.  We assume optical imaging dimensions are $2 N_{x} \times 2 N_{y} = 1024 \times 1024$. We use a wavelength $\lambda = 532e-6$ \ nm and optical wavelength number $k = (2 \cdot \pi)/\lambda = 11,810.50$.  

We run various non-convex CGH optimization algorithms for 3D holograms of a volumetric reconstructed brain from MRI slices, an image of a 3D brain model, and fluorescent cell image in microscopy to compare their reconstruction performance.   The flexibility of the optimization frameworks also brings with it a diversity of pipelines for hologram synthesis of 3D objects. Many diffractive propagation models are applicable for volumetric optimization. For a better illustration of different optimization pipelines, they are unified into the band-limited ASM.  We use the superposition and global methods for alternating projections (AP), stochastic gradient descent (SGD), and quasi-newton (QN) method. We also use the sequential method for alternating projections.  We generate both complex Fourier transform (FFT) holograms and phase-only holograms (POH) for each method.

\begin{figure}[h!]
	\centering
	\begin{center} 
	\includegraphics[width=0.7\linewidth]{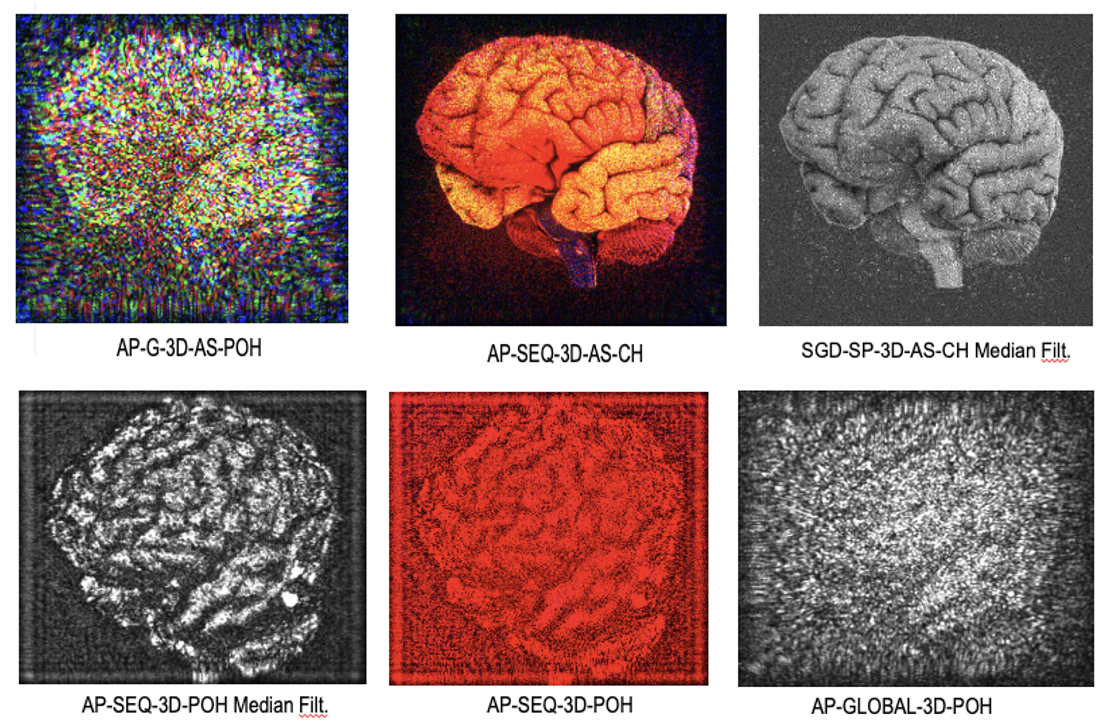}
	%\small \caption{} 	
	\caption{Snap Shot During Method Optimizations. Top row: 3D Model brain. Bottom row: Volume reconstructed brain} 
	\label{fig:recon2}  
	\end{center}  
\end{figure}

Figures \ref{fig:panel2} and \ref{fig:im1} show a visualization of the various parameters used in the Fourier optics optimization and computation process for FFTs and IFFTs of the hologram synthesis for the fluorescent cell and 3D brain model, respectively, including amplitude, phase, point source function (PSF), and transfer function $H$.  Figure \ref{fig:im1} illustrates the reconstructed holograms for various optimization methods.  Figure \ref{fig:recon2} shows a snap shot during one iteration of the reconstruction optimization process.

%\begin{table}
%\begin{center}
%\begin{tabular}{|c c c c c c c|} 
% \hline
% Optim. & CH/POH & 2D/3D & AS/FFT & MSE & RMSE & PSNR & MSE_{\med} & PSNR_{\med.} & depth \\ 
% \hline
% %SGD & CH & 2D & AS \\ 
   
\begin{figure}[h!]
	\centering
	\begin{center} 
	\includegraphics[width=0.8\linewidth]{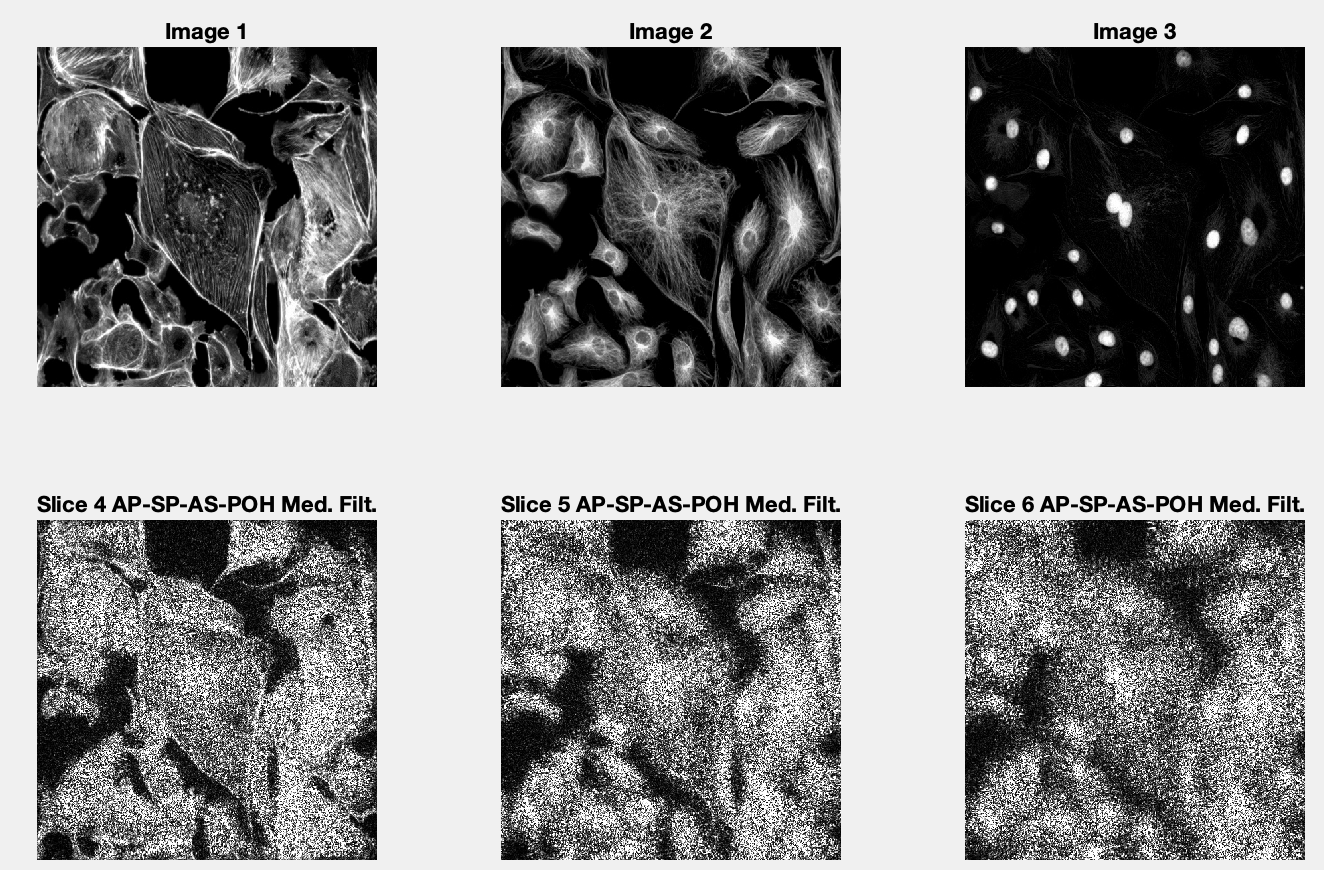}
	%\small \caption{} 	
	\caption{Optimized Cell Hologram Slices} 
	\label{fig:cell4}  
	\end{center}  
\end{figure}
\begin{figure}[h!]
	\centering
	\begin{center} 
	\includegraphics[width=0.8\linewidth]{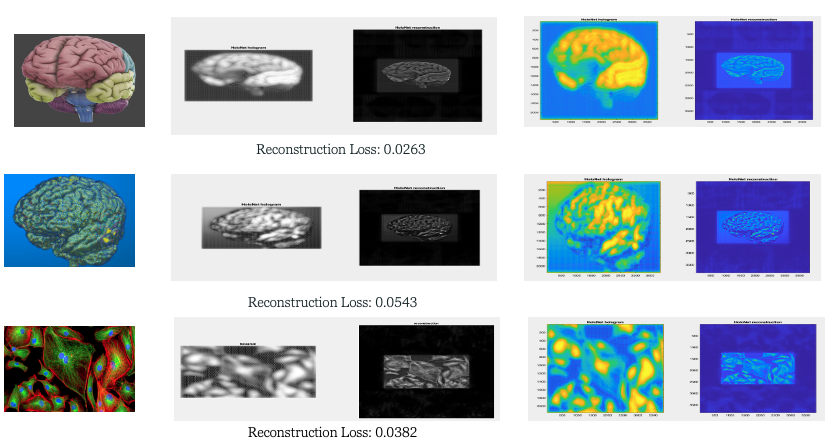}
	%\small \caption{} 	
	\caption{Predicted Holograms Using HoloNet.  Top row: 3D brain model. Middle row: volumetric brain.  Bottom row: fluorescent cell} 
	\label{fig:dl}  
	\end{center}  
\end{figure}

 
    Figure \ref{fig:dl} shows the predicted reconstruction images and losses using  HoloNet.  HoloNet was trained on 800 DIV2K images and validated on 100 DIV2K images using 10 epochs, 2000 iterations, a learning rate of 0.001, and gradient decay factor of 0.9.  As shown, the volumetric brain had a reconstruction MSE loss of 0.0543, the 3D brain model image had a reconstructon MSE loss of 0.0263, and the fluorescent cell image had a reconstruction MSE of 0.0382.
    Table \ref{table:tbl1} shows the optimized results for all optimization algorithms for the volumetric brain reconstruction from point cloud data.  The algorithms were run both with and without 2D median filtering during the optimization iteration reconstruction process.  The PSNR that is highest with or without filtering is highlighted light gray. 
    As shown, the superposition (SP) 3D-AS-CH method has the highest PSNR of 44.08 with filtering of all the methods.   In only 4 out of 14 cases was the PSNR higher without filtering.  In fact, for certain algorithms, 2D median filtering can substantially reduce MSE and increase PSNR.  For instance, in the SGD SP-3D-AS-POH method increased PSNR by $36.6\%$ while in the quasi-Newton SP-3D-AS-POH method, it increased PSNR by $78.9\%$ while reducing MSE by $40.4\%$.
%\begin{table*}[t]
%\small
%\begin{center}
%\begin{tabular}{|c c c c c c c|} 
%\hline
%Optim. & CH/POH & 2D/3D & AS/FFT & MSE & RMSE & PSNR  \\ 
%\hline
 %SGD & CH & 2D & AS \\ 
%AP-SP & POH & 3D & AS & 0.0012 & 0.0346 & 29.22  
%AP-S & POH & 3D & AS & 1.2795 & 1.1311 & -1.07  
%AP-SP & CH & 3D & AS & 0.0011 & 0.0333 & 39.81  
%QN  & POH & 2D & FFT & 0.2165 & 0.4653 & 6.64  
%QN-SP & CH & 3D & AS & 0.0114 &  0.1067 & 29.30  
%QN-SP &  POH & 3D & AS & 0.0649 & 0.2548 & 11.88 \\   
%QN-G & POH & 3D & AS & 0.0204 & 0.1428 & 26.81 \\
%QN-G & CH & 3D & AS & 0.0750 & 0.2738 & 11.27 \\
%SGD & CH  & 3D & AS & 1.0514 & 1.0254 &    \\ %[823.7, 564.0, 133.5] \\
%SGD & POH & 2D & FFT & 0.1738 & 0.4169 & \\ 
%SGD & POH & 2D & AS  & 1.558 & 1.248 & -1.926\\
% SGD & CH & 3D & AS & 0.1025 & 0.3102 & & \\
%SGD & POH & 3D & AS & 8.24e-4 & 0.0287 & 30.84 \\
%SGD & CH & 3D & AS & 8.18e-4 & 0.0286 & 41.03 \\
 %\hline
%\end{tabular}
%\end{center}
%\end{table*}

 %AP-SP & POH & 3D & AS & 0.0052 &0.0722 & 22.83 &       & 80 \\
 %AP-G & CH & 3D & AS & 0.0942 & 0.3069 & 20.63 &         & 22 \\
 %AP-G & POH & 3D & AS & 0.1261 & 0.3551 & 8.99 & 80 \\
 %QN-SP & POH & 3D & AS & 0.0735 & 0.2712 & 11.34 & 22

\begin{figure}[h!]
	\centering
	\begin{center} 
	\includegraphics[width=1\linewidth]{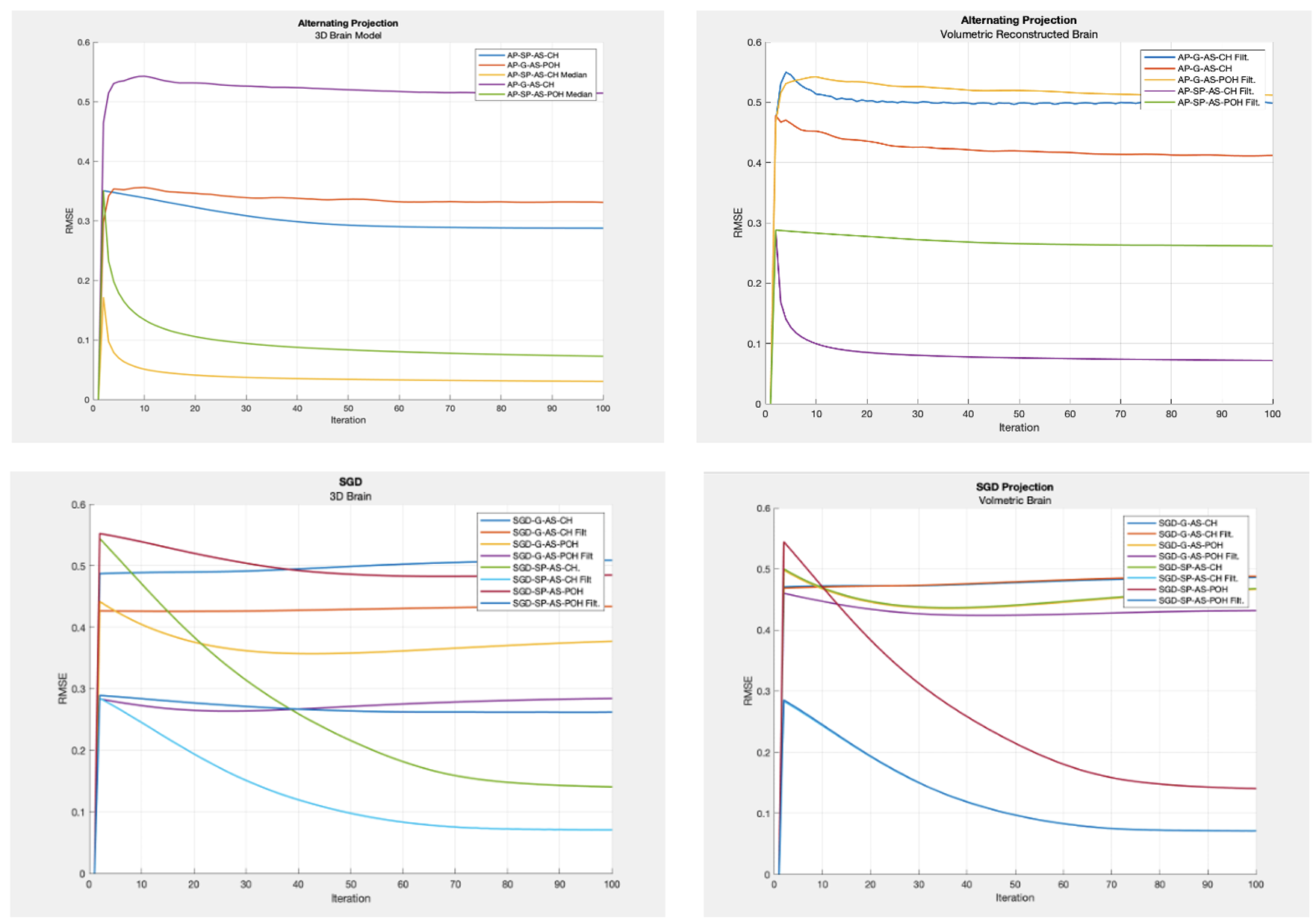}
	%\small \caption{} 	
	\caption{RMSE Comparison of (a) Alternating Projection and (b) SGD Methods} 
	\label{fig:sgd10}  
	\end{center}  
\end{figure}

\begin{table*}[h!] 
	{
		\scriptsize	
		%\caption{Comparison of minimum path length, average path length of proposed Advised-RRT*, Informed RRT* and RRT* in a map size 1073$\times$1073.} 
		\vspace{-0.4cm}
		\begin{center}
        \caption{CGH Optimization Comparison from Volumetric Brain Reconstruction}
         \label{table:tbl1}
			\begin{tabular}{ c  c c  c  c    c  c  c  c}
                \hline 
                %\hline
                &   &  &     \textbf{w/o Filt.}  &  &  &    \textbf{w/ Filt.} \\
                &   Model & Depth & MSE & RMSE & PSNR &  MSE & RMSE & PSNR \\  
				%\hline \hline 
				%\rm \tabincell{c}{Map \\ Name} & 
				%\rm \tabincell{c}{Model \\ name} & 
				%\rm \tabincell{c}{Minimum \\ path length ($m$)} &
				%\rm \tabincell{c}{Average \\ path length ($m$)} 
				\\
				\hline
				 & SP-3D-AS-CH & 22 & 0.0052 & 0.0724 & 35.86 & 0.009 & 0.0304 & \cellcolor{black!10} 44.08 \\
				%\cline{2-8}
\multirow{4}{*}{Alternative Projection} & SP-3D-AS-POH & 22 & 0.0053 & 0.0725 & 22.79 & 0.0009 & 0.0304 & \cellcolor{black!10} 30.35  \\
				%\cline{2-8}
%& SP-3D-AS-CH & 80 & 0.0957 & 0.3093 & 20.93 & 0.1064 & 0.3262 & 20.89 \\
& SEQ-3D-AS-CH & 10 & 0.3917 & 0.6259 & 4.27 & 0.150 & 0.3873 & \cellcolor{black!10} 8.43  \\
& SEQ-3D-AS-POH & 10 & 0.2706 &0.5202 & 5.87 & 0.1051 & 0.3242 & \cellcolor{black!10} 9.98 \\
& G-3D-AS-CH & 80 & 0.0942 & 0.3069 & \cellcolor{black!10} 20.63  &  0.2481 & 0.4981 & 17.31 \\
& G-3D-AS-POH & 80 & 0.1261 & 0.3551 & \cellcolor{black!10} 8.99  & 0.1096 & 0.3310 & 9.60 \\
				\hline
\multirow{4}{*}{Quasi-Newton} 
& G-3D-AS-CH & 40 & 0.1604 & 0.4006 & 7.95 & 0.1594  & 0.3993 & \cellcolor{black!10} 7.98 \\
				%\cline{2-8}
& G-3D-AS-POH & 40  & 0.2093 & 0.4575 & 6.79 & 0.0894 & 0.299 & \cellcolor{black!10} 10.49  \\
				%\cline{2-8}
& SP-3D-AS-CH & 22 & 0.0826 & 0.2874 & 23.63 & 0.0229 & 0.1512 & \cellcolor{black!10} 29.20 \\
				%\cline{2-8}
& SP-3D-AS-POH & 22  & 0.0740 & 0.2721 & \cellcolor{black!10} 24.29  & 0.0227 & 0.1508 & 16.43 \\
\hline
\multirow{4}{*}{SGD}  
& G-3D-AS-CH & 70 & 0.2595 & 0.5094 & 5.95  & 0.0637 & 0.2510 & \cellcolor{black!10} 12.1 \\
				%\cline{2-8}
& G-3D-AS_POH & 70 & 0.1261  & 0.3551 & 8.99  & 0.0829  & 0.2879  & \cellcolor{black!10} 10.82 \\
& SP-3D-AS-CH & 22 & 0.0057 & 0.0757 & \cellcolor{black!10} 35.4 & 0.0231 & 0.1520 & 29.43  \\ %& 53706
				%\cline{2-8}
& SP-3D-AS-POH & 22 & 0.0072 & 0.0847 & 21.44 & 0.0012 & 0.0347 &  \cellcolor{black!10} 29.20 \\
            
%& SP-3D-AS-POH & 22 & 0.0730 & 0.2702 & 11.37 &  & 0.0722 & 22.83  \\
				%\cline{2-8}
            \hline 
			\end{tabular}
		\end{center}
	}
        
\end{table*} 
  Table \ref{table:tbl2} shows the optimized results for each CGH opitmization method for the image of the 3D brain model.  Interestingly, the MSE and PSNR are overall lower than in the reconstructed volumetric brain case.  However, the results are consistent that 2D median filtering can improve MSE, RMSE, and PSNR in most algorithms.
\begin{table*}[h!] 
	{
		\scriptsize	
		%\caption{Comparison of minimum path length, average path length of proposed Advised-RRT*, Informed RRT* and RRT* in a map size 1073$\times$1073.} 
		\vspace{-0.4cm}
		\begin{center}
            \caption{CGH Optimization Comparison from Image of 3D Brain Model}
            \label{table:tbl2}
			\begin{tabular}{ c  c c  c  c    c  c  c  c}
                \hline 
                \hline
                &   &  &     \textbf{w/o Filt.}  &  &  &    \textbf{w/ Filt.} \\
                &   Model & Depth & MSE & RMSE & PSNR &  MSE & RMSE & PSNR \\  
				%\hline \hline 
				%\rm \tabincell{c}{Map \\ Name} & 
				%\rm \tabincell{c}{Model \\ name} & 
				%\rm \tabincell{c}{Minimum \\ path length ($m$)} &
				%\rm \tabincell{c}{Average \\ path length ($m$)} 
				\\
				\hline
				 & SP-3D-AS-CH & 22 & 0.0176  & 0.1327 & 30.1 & 0.0051 & 0.0717 & \cellcolor{black!10} 35.19 \\
				%\cline{2-8}
\multirow{4}{*}{Alternative Projection} & SP-3D-AS-POH & 22 & 0.0177 & 0.1329 & 17.5 & 0.0051 & 0.0715 & \cellcolor{black!10} 22.91  \\
				%\cline{2-8}
& SEQ-3D-AS-CH & 10 & 1.144 & 1.07 & -0.646 & 0.3797 & 0.6162 & \cellcolor{black!10} 4.14 \\
& SEQ-3D-AS-POH & 10 & 0.8179 & 0.9044 & 0.8105 & 0.2569 & 0.5068 & \cellcolor{black!10} 5.84 \\
%& G-3D-AS-CH & 80 & 0.8208  & 0.9057 & 0.7978 & 0.2569 & 0.5068 & 5.84  \\
& G-3D-AS-CH & 80 & 0.1312 & 0.3622 & \cellcolor{black!10} 19.34  & 0.2470 & 0.4970 & 16.86  \\
& G-3D-AS-POH & 80 & 0.1695 & 0.4116 & \cellcolor{black!10} 7.71 & 0.2644 & 0.5142 & 5.77   \\
				\hline
\multirow{4}{*}{Quasi-Newton} & G-3D-AS-CH & 40 & 0.1673 & 0.4091 & \cellcolor{black!10} 7.739 & 0.1941 & 0.4406 & 7.095    \\
				%\cline{2-8}
& G-3D-AS-POH & 40  & 0.2286  & 0.4781  & 6.41  & 0.2179 & 0.4668 & \cellcolor{black!10} 6.62   \\
				%\cline{2-8}
& SP-3D-AS-CH & 22 & 0.0745  & 0.2730 & 11.28 & 0.0734 & 0.2709 & \cellcolor{black!10} 11.34     \\
				%\cline{2-8}
& SP-3D-AS-POH & 22 & 0.2072  & 0.4552 & 6.84 & 0.0734 & 0.2709 & \cellcolor{black!10} 11.34  \\
				\hline
                \multirow{4}{*}{SGD}  & G-3D-AS-CH & 70 & 0.1673 & 0.4091 & \cellcolor{black!10} 7.739 & 0.1941 & 0.4406 & 7.095    \\
				%\cline{2-8}
& G-3D-AS_POH & 70 & 0.2202 & 0.4692 & 6.57 & 0.1853 & 0.4304 & \cellcolor{black!10} 7.32    \\
& SP-3D-AS-CH & 22 & 0.0197 & 0.1404 & 30.5 & 0.0051 & 0.0711 & \cellcolor{black!10} 37.31       \\
				%\cline{2-8}
& SP-3D-AS-POH & 22  & 0.0195 & 0.1398 & 17.09 & 0.005 & 0.0708 & \cellcolor{black!10} 23    \\
				%\cline{2-8}
            \hline 
			\end{tabular}
		\end{center}
	}
\end{table*}
Figure \ref{fig:sgd10} plots RMSE curves of different alternating projection methods for the 3D model brain and point cloud brain.  As shown, 2D median filtering substantially improves reconstruction performance and provides the lowest errors during optimization.  Figure \ref{fig:sgd10} also plots the RMSE curves of SGD methods.  As shown, SGD median filtering also reduces RMSE but is not as pronounced as in the AP method.
\section{3D Hologram Projection}

\begin{figure}[h!]
	\centering
	\begin{center} 
	\includegraphics[width=0.8\linewidth]{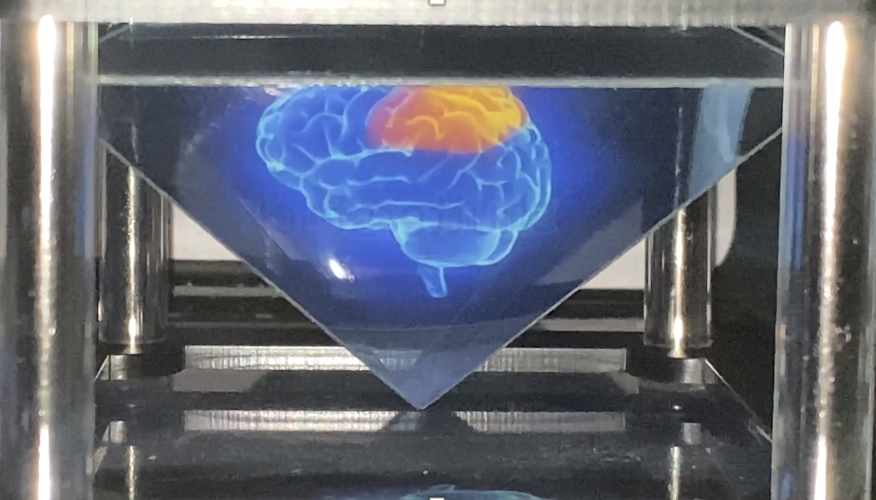}
	%\small \caption{} 	
	\caption{3D Display using Pepper's Ghost Effect} 
	\label{fig:project}  
	\end{center}  
\end{figure}

    After the hologram is generated, it needs to be displayed with a SLM or 3D projector.  There are various holographic methods for augmented reality (AR) and virtual reality (VR) displays such as foveated displays. \cite{Park:2022}.  Since we have simulated a SLM and do not have a physical SLM for holographic projection, we use Pepper's ghost effect instead to project a 3D brain in a prism as illustrated in Figure \ref{fig:project}.   
\section{Conclusion}
    An efficient and fast pipeline was proposed to synthesize 3D holograms from point cloud and scanned MRI data. Fourier optics using non-convex Optimization is an effective method for CGH using point clouds and overall performed better than using a 3D model  image. Deep learning for CGH like HoloNet can improve reconstruction loss but require GPUs and a substantial amount of time and computing resources. Our results suggest that optimization methods are faster and can outperform deep learning in certain cases.    For instance, the alternating projection method with angular spectrum for CH provides the highest PSNR. Moreover, 2D Median filter sharpening can substantially improve PSNR and reduce MSE during optimization and reconstruction by removing artifacts and speckled noise. 
    
    The tradeoff between PSNR and hologram depth is approximately inverse.   Future research in field experiments using an actual spatial light modulator and pulse laser hardware and applications for real-time reconstruction need to be conducted.  Future research will include compressive sensing in the analysis for use in real-time CGH, compare the performance with different point spread function (PSF) shapes, and perform 3D projection with an a physical SLM. 

- 
%\end{itemize}
%\textbf{Week 8: 11/1/2024 - 11/7/2024}
%\begin{itemize}
%\item Implementation and compare different algorithms. Start training.
%\end{itemize}
%\textbf{Week 9: 11/8/2021 - 11/14/2021} 
%\begin{itemize}
%\item Analyze preliminary results. Revise approach and conduct ablative study if necessary.
%\end{itemize}
%\textbf{Week 10: 11/15/2021 - 11/19/2024}
%\begin{itemize}
%\item Analyze and summarize the results.
%\item Submit Report and code. (12/19/2021, 11:59 pm)
%\end{itemize}
\printbibliography
%\bibliographystyle{IEEEtran}
%\bibliography{biblio} 

\end{document}